%% file: acl_latex.tex
\documentclass[11pt]{article}

\usepackage[final]{acl}

\usepackage{times}
\usepackage{latexsym}

\usepackage[T1]{fontenc}

\usepackage[utf8]{inputenc}

\usepackage{tikz}
\usepackage{graphicx}
\usetikzlibrary{arrows.meta,shadings}

\usepackage{microtype}

\usepackage{inconsolata}

\usepackage{graphicx}

\usepackage{subcaption}

\usepackage{array}
\usepackage{booktabs}
\usepackage{algorithm}
\usepackage{algorithmic}

    \title{Iterative Topic Taxonomy Induction with LLMs: A Case Study of Electoral Advertising}

\author{Alexander Brady\thanks{Equal contribution.} \\
  Department of Computer Science  \\
  ETH Zurich \\
   Zurich, Switzerland \\
  \texttt{bradya@ethz.ch} \\\And
   Tunazzina Islam\footnotemark[1]  \\
  Department of Computer Science  \\
  Purdue University \\
  West Lafayette, IN 47907, USA \\
  \texttt{islam32@purdue.edu} \\}

\begin{document}
\maketitle

\begin{abstract}
Social media platforms play a pivotal role in shaping political discourse, but the scale and rapid evolution of online content make systematic analysis difficult. We introduce an end-to-end framework for inducing an interpretable topic taxonomy from unlabeled text corpora. The framework combines embedding-based clustering with iterative large language model (LLM) inference to construct a topic taxonomy without requiring predefined labels or seed topics. It first synthesizes candidate topics from document clusters and then uses the resulting taxonomy to assign consistent topic labels across clusters. We evaluate the approach through a case study of political advertising ahead of the 2024 U.S. presidential election. We use the induced taxonomy to support downstream analyses of issue prevalence, moral framing, advertising spend, and demographic exposure patterns. These results suggest that iterative taxonomy construction can provide a scalable and interpretable approach to organizing large unlabeled text corpora while supporting substantive downstream analysis.
\end{abstract}

\begin{figure}[t]
    \centering
    \resizebox{\linewidth}{!}{%
        \input{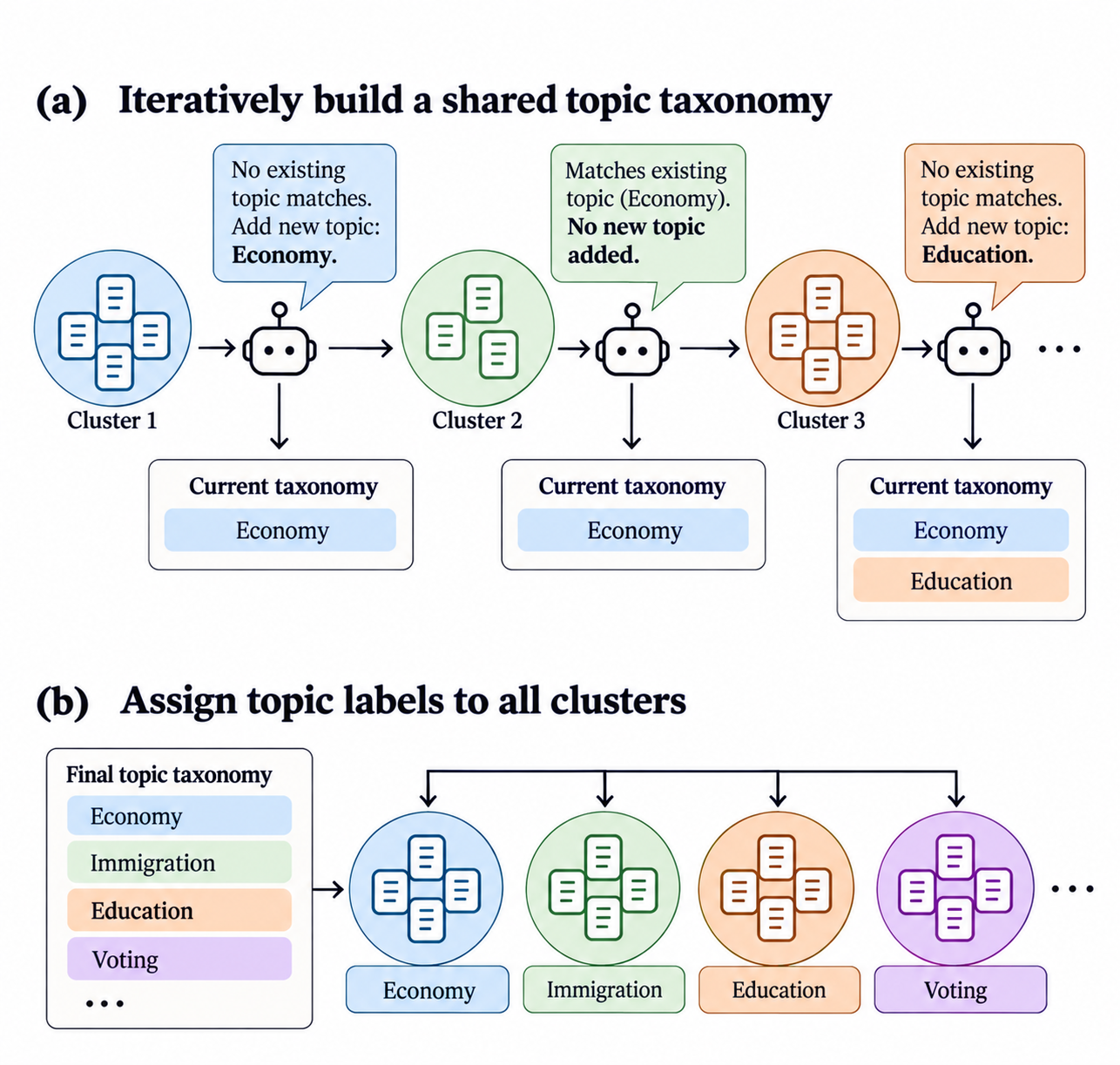}%
    }
    \caption{\small {Overview of the proposed framework. Documents are first embedded and clustered. (a) Cluster representatives are processed sequentially by the LLM to induce a shared topic taxonomy. (b) The LLM assigns one topic label to each cluster from the completed taxonomy.}}
    \label{fig:overview}
\end{figure}

\section{Introduction}
Modern political campaigns have been fundamentally transformed by the rise of social media platforms. Platforms such as Facebook and Instagram have become important venues for political advertising, enabling campaigns and advocacy organizations to reach voters with issue-specific appeals. Prior work has shown that voters respond more favorably to advertisements addressing issues they personally prioritize, irrespective of partisanship~\citep{chu2024or}. Understanding the content and structure of these messages is critical for studying contemporary political communication.

The scale and rapid evolution of social media content make manual analysis difficult. Traditional topic modeling methods such as Latent Dirichlet Allocation (LDA) \cite{blei2003latent} and non-negative matrix factorization (NMF) \cite{lee1999learning} provide unsupervised mechanisms for discovering latent structure in text corpora, but their outputs often require additional interpretation or manual labeling. Prior work has therefore explored automated topic labeling through probabilistic methods \cite{magatti2009automatic}, summarization-based approaches \cite{wan2016automatic,cano2014automatic}, and sequence-to-sequence models \cite{alokaili2020automatic}.

More recently, embedding-based approaches such as BERTopic \cite{grootendorst2022bertopic} have combined contextualized text representations with density-based clustering to produce more coherent topic groupings. Large language models (LLMs) have also enabled more semantically expressive approaches to topic discovery and labeling \cite{islam-2026-reasoning,sudhoff2026paid,pham2024topicgpt}. However, several recent LLM-based approaches assume either an initial seed set of themes or a predefined set of topics~\citep{islam2025discovering,islam-goldwasser-2025-uncovering}.

In this work, we introduce a seed-free framework for inducing an interpretable topic taxonomy from unlabeled text corpora (Figure~\ref{fig:overview}). Our method first uses embedding-based clustering to identify groups of semantically related documents. It then processes these clusters sequentially with an LLM, which determines whether each cluster is adequately represented by an existing topic or whether a new topic should be added to the taxonomy. In a second pass, the complete induced taxonomy is used to assign a single topic label to each cluster. This iterative procedure is designed to promote consistency across cluster labels while allowing the taxonomy to expand as new themes are encountered.

Unlike methods that require human-provided seed labels, our framework constructs the taxonomy directly from the corpus. The method is also distinct from single-shot LLM labeling: rather than generating each cluster label independently, previously induced topics remain available throughout the synthesis process, creating a shared and evolving label space across clusters. The resulting taxonomy can then be used for downstream annotation and analysis.

We evaluate the framework through a case study of political advertisements collected from the Meta Ad Library during the month preceding the 2024 U.S. presidential election\footnote{Code, prompts and dataset are available at: \url{https://github.com/alexander-brady/llm-topic-synthesis}}. The induced taxonomy captures a range of political issues represented in the advertising corpus. We further extend our analysis by examining the moral framings underlying these ads through the lens of Moral Foundations Theory~\cite{iyer2012understanding,haidt2007morality,haidt2004intuitive}. 
By examining political issue framing and its associations with observed audience demographics, the case study illustrates the substantive analyses enabled by the proposed framework.

Our contributions are:
\begin{enumerate}\setlength{\itemsep}{0pt}\setlength{\parskip}{0pt}
    \item A general, seed-free, LLM-guided framework for inducing interpretable topic taxonomies from unlabeled corpora without human intervention.
    \item An iterative topic synthesis mechanism that yields more consistent and better-aligned topic labels than classical topic models and single-shot LLM labeling under human evaluation.
    \item A case study of 2024 U.S. political advertising demonstrating how the induced taxonomy can support downstream analyses of issue salience, moral framing, ad spending, and demographic exposure.
    \item Release of a curated dataset of 2024 U.S. presidential election ads to support future research.
\end{enumerate}

\section{Related Work}

\paragraph{Political advertising and microtargeting.}
Social media has become an important venue for political communication, campaign messaging, and voter mobilization~\citep{islam2025understanding,aggarwal20232,capozzi2021clandestino,goldberg2021shifting,ribeiro2019microtargeting,silva2020facebook,fulgoni2016power,teresi2015wired,hersh2015hacking}. A substantial body of work has examined political microtargeting and the ways campaigns tailor messages to specific audiences~\citep{islam2025post,islam2025understandingthesis,tappin2023quantifying,zuiderveen2018online,hersh2015hacking}. 

Related studies have analyzed the strategic dissemination of campaign content across platforms and the relationship between targeted advertising and voter responses~\cite{hackenburg2024evaluating,islam2023weakly,hirsch2024beneficial}. These settings motivate scalable methods for organizing and interpreting large collections of political messages.

\paragraph{Topic modeling and automated topic labeling.}
Traditional topic modeling methods have been widely used to analyze large text corpora in domains including social media \cite{chen2019experimental,ozer2016community} and political discourse \cite{mathaisel2021political,lahoti2018joint}. However, classical approaches can produce topics that are difficult to interpret and often require human inspection or post-hoc labeling~\cite{chang2009reading,mei2007automatic}. More recent embedding-based approaches such as BERTopic \cite{grootendorst2022bertopic} combine contextualized representations with density-based clustering to improve topic coherence and interpretability.

Large language models have enabled a further shift toward direct topic synthesis and labeling. TopicGPT \cite{pham2024topicgpt}, for example, uses LLMs to construct and refine topic taxonomies and has demonstrated strong performance in terms of coherence and interpretability. More broadly, LLMs have also shown strong performance on annotation tasks traditionally performed by human coders \cite{gilardi2023chatgpt}. These results motivate the use of LLM inference not only for labeling existing clusters but also for constructing a shared topic inventory.

Recent LLMs-in-the-loop methods have explored fine-grained theme and argument discovery, for example using an initial seed set of themes to guide topic discovery or uncover associated arguments~\citep{islam-goldwasser-2025-uncovering,islam2025discovering,islam2023analysis,islam2022understanding}. \citet{lam2024concept} introduced LLooM, which iteratively induces and scores higher-level concepts from text, with optional analyst-provided seeds. In contrast, our framework incrementally constructs a shared topic inventory over document clusters and then relabels all clusters using the completed taxonomy.

\paragraph{Moral framing in political discourse.}
Moral Foundations Theory provides a widely used framework for analyzing the moral dimensions of political language \cite{iyer2012understanding,haidt2007morality,haidt2004intuitive}. Earlier computational approaches typically relied on manually annotated datasets to train classifiers for detecting moral appeals~\cite{pacheco2022holistic,roy2021identifying}. More recently, LLM-based approaches have been used to generate moral labels and explanations directly from political text \cite{islam2025can}. We build on this line of work by using the induced topic taxonomy as the basis for a downstream analysis of moral framing in political advertisements. In our setting, moral annotations are used as an exploratory analytical layer rather than as independent validation of the induced topics.

\section{Methodology}
\label{sec:methodology}

We propose a framework for inducing an interpretable topic taxonomy from an unlabeled text corpus. This framework consists of three main stages: (1) embedding-based clustering, (2) LLM-based taxonomy induction, (3) cluster-level topic assignment. We formalize the iterative topic synthesis and annotation procedure in Algorithm~\ref{alg:topic_taxonomy}.

\subsection{Embedding-Based Clustering}

Given a corpus of documents, we first encode each document into a dense vector representation using a pretrained sentence embedding model such as Sentence-BERT \citep{reimers2019sentence}. We then apply UMAP \citep{mcinnes2020umapuniformmanifoldapproximation} to reduce the dimensionality of the embedding space before clustering.

We cluster the reduced embeddings using HDBSCAN \cite{mcinnes2017hdbscan}, a density-based clustering algorithm that can identify clusters of varying density while leaving low-confidence points unassigned. This property is useful for noisy text corpora in which not every document necessarily belongs to a well-defined topical cluster.

For each cluster $C_i$, we select the five documents with the highest HDBSCAN membership probabilities as its representative set $R(C_i)$. These representative documents are used as the textual input to the LLM during both taxonomy induction and subsequent topic assignment. Restricting LLM inference to a small representative set reduces the amount of text that must be processed while emphasizing documents that are most characteristic of each cluster.

\begin{algorithm}[t]
\caption{Iterative Topic Taxonomy Induction}
\label{alg:topic_taxonomy}
\begin{algorithmic}[1]
\REQUIRE Document clusters $\mathcal{C} = \{C_1, \dots, C_n\}$, 
cluster representatives $R(C_i)$, 
LLM $\mathcal{M}$
\ENSURE Topic taxonomy $\mathcal{T}$ and cluster-topic assignments

\STATE Initialize topic list $\mathcal{T} \leftarrow \emptyset$

\FOR{each cluster $C_i \in \mathcal{C}$}
    \IF{$\mathcal{T} = \emptyset$}
        \STATE Prompt $\mathcal{M}$ to generate a topic $t$ from $R(C_i)$
        \STATE $\mathcal{T} \leftarrow \mathcal{T} \cup \{t\}$
    \ELSE
        \STATE Prompt $\mathcal{M}$ with $R(C_i)$ and topics $\mathcal{T}$
        \STATE Constrain output to \{\texttt{yes}, \texttt{no}\}
        \IF{\texttt{no}}
            \STATE Prompt $\mathcal{M}$ to generate a new topic $t$ from $R(C_i)$
            \STATE $\mathcal{T} \leftarrow \mathcal{T} \cup \{t\}$
        \ENDIF
    \ENDIF
\ENDFOR

\FOR{each cluster $C_i \in \mathcal{C}$}
    \STATE Prompt $\mathcal{M}$ with $R(C_i)$ and topics $\mathcal{T}$
    \STATE Constrain output to select one topic $t \in \mathcal{T}$
    \STATE Assign topic $t$ to cluster $C_i$
\ENDFOR

\RETURN $\mathcal{T}$ and cluster-topic assignments
\end{algorithmic}
\end{algorithm}

\subsection{Iterative Topic Taxonomy Induction}

The goal of the next stage is to construct a shared topic inventory directly from the document clusters, without requiring human-provided seed labels. We initialize the topic taxonomy $\mathcal{T}$ as an empty set and process the clusters sequentially.

For the first cluster, the LLM is prompted to generate a concise topic label from its representative documents. This label becomes the first element of $\mathcal{T}$. For each subsequent cluster, the LLM is shown the cluster representatives together with the topics currently in $\mathcal{T}$ and asked whether any existing topic adequately describes the cluster. We use constrained decoding \cite{beurer2024guiding} to restrict this decision to \texttt{yes} or \texttt{no}. If the model returns \texttt{no}, it is then prompted to generate a new concise topic label, which is then added to $\mathcal T$.

In this way, previously generated labels remain available while later clusters are processed. Unlike independent, single-shot labeling, the procedure therefore constructs an evolving shared label space and only introduces a new topic when the model determines that the existing inventory is insufficient.

For our experiments, clusters were processed in ascending HDBSCAN label order. Because HDBSCAN cluster identifiers are arbitrary rather than semantically ordered, this corresponds to a fixed but otherwise arbitrary traversal of the induced clusters. We treat robustness to cluster ordering as a limitation and direction for future work.

\subsection{Cluster-Level Topic Assignment}

The first pass determines which topics belong in the taxonomy but does not necessarily provide the final assignment for every cluster. After the complete topic inventory has been constructed, we therefore perform a second pass over all clusters.

For each cluster, the LLM receives its representative documents together with the complete taxonomy $\mathcal{T}$. We use constrained decoding to require the model to select exactly one label from the available topic inventory. This second pass allows every cluster to be evaluated against the same final label space, including topics that may have been introduced only after that cluster was first encountered. The output of the framework is therefore (1) an induced topic taxonomy and (2) a topic assignment for each document cluster. 



\section{Case Study: Experimental Setup}

We apply the proposed framework to political advertisements from the 2024 U.S. election cycle. The goal of the case study is twofold: first, to evaluate the quality of the induced topic taxonomy; and second, to demonstrate how the resulting annotations can support downstream analyses.

\subsection{Dataset}
\label{sec:case-study-dataset}

We collected political advertisements running on Facebook and Instagram in the United States during October 2024 using the Meta Ad Library, starting one month before the 2024 presidential election. The resulting dataset contains $8{,}047$ unique advertisements. The ads cover election-related messaging at both the local and federal levels. $99\%$ of the advertisements are in English, with a small fraction in Spanish.

For each advertisement, the Meta Ad Library provides a unique ad identifier, title, body text, URL, creation time, campaign duration, page author, funding entity, and advertising spend reported as a range. We represent spend using the midpoint of the reported range. Meta also reports impressions as a range, for which we similarly use the midpoint, together with aggregate audience distributions by gender, age group, and U.S. state.

Because these demographic statistics describe observed ad delivery rather than necessarily the advertiser's intended targeting criteria, we refer to them throughout as \textit{audience exposure} or \textit{delivery patterns} rather than direct targeting decisions.

\subsection{Taxonomy Induction}

We embed the advertisement body texts using the \texttt{all-MiniLM-L6-v2} sentence-transformer model. To reduce near-duplicate content, we remove advertisements whose embedding has cosine similarity greater than $0.95$ to another advertisement. We then apply HDBSCAN to the UMAP-reduced embeddings with a minimum cluster size of $15$.

This procedure assigns $4{,}510$ advertisements to $73$ clusters. For each cluster, we select the five advertisements with the highest HDBSCAN membership probabilities as cluster representatives. These are given to the LLM during taxonomy induction and cluster-level topic assignment.

We use a locally hosted Llama-3.2-3B model~\cite{grattafiori2024llama} to execute our framework, without GPU acceleration.

We initialize the topic inventory with no seed labels. Processing the $73$ clusters using the procedure described in Section~\ref{sec:methodology} produces a taxonomy containing $15$ generated topics. During the second-pass cluster assignment stage, the model is additionally allowed to assign the label \textit{other} when none of the induced topics adequately describes a cluster.
Although \textit{border security} appears in the generated taxonomy, no cluster is ultimately assigned this label.
The final topic set and cluster counts are shown in Table~\ref{tab:topics}. The complete prompts are provided in App. ~\ref{app:prompts}.

\begin{table*}
    \centering
    \small
    \begin{tabular}{lc>{\raggedright\arraybackslash}p{10cm}}
        \hline
        \textbf{Issue} & \textbf{\#Clusters} & \textbf{Example Ad} \\
        \hline
        economy & 15 & \textit{Molly Buck's agenda is failing Iowa families. We can't afford Molly Buck in the State House.} \\
        \hline
        voting rights & 13 & \textit{Make Your Vote Count. RE-register to Vote in the District of Your Second Home.} \\
        \hline
        crime/justice & 12 & \textit{Orange County Firefighters Trust Dave Min To Keep Our Communities Safe.} \\
        \hline
        personal freedom & 4 & \textit{Tim Sheehy will always fight for Montana Values in the Senate!} \\
        \hline
        voting & 4 & \textit{Vote Rebecca for State Rep by Nov 5th!} \\
        \hline
        education & 4 & \textit{Vote for students. Vote for teachers. Support our public schools.} \\
        \hline
        affordable housing & 4 & \textit{Catherine has supported more multi-family housing so we can afford homes in the communities we grew up in. Vote Stefani!} \\
        \hline
        environmental protection & 4 & \textit{We can count on Lucy Rehm to be a responsible leader and protect our clean air and water.} \\
        \hline
        immigration & 3 & \textit{Yadira Caraveo supported efforts to defund ICE and the border patrol. Now Colorado is being overwhelmed with illegal immigration.} \\
        \hline
        election integrity & 3 & \textit{Question 3 will change our state's Constitution and weaken our democracy by throwing away thousands of ballots over minor errors. Vote NO on Question 3.} \\
        \hline
        abortion & 2 & \textit{Defend reproductive freedom - VOTE on Tuesday, November 5th!} \\
        \hline
        healthcare access & 1 & \textit{Lucy Rehm cares about improving conditions in your local hospital. That's why Minnesota Nurses endorsed Lucy Rehm!} \\
        \hline
        property taxes & 1 & \textit{Caleb will work to stop reckless spending and oppose tax and fee increases.} \\
        \hline
        sewer sale & 1 & \textit{Get the facts on the Gloucester Township sewer sale, visit VoteYesGloucesterTownship.com!} \\
        \hline
        other & 2 & \textit{LISCOMB - thanks for welcoming me at the little street where old and new friends meet!} \\
        \hline
    \end{tabular}
    \caption{{\small Topics identified by the LLM in the political ads dataset. `\#Clusters' column gives the number of unique clusters assigned to each topic by the LLM.}}
    \label{tab:topics}
\end{table*}

\subsection{Human Evaluation: Topic Labels}

To evaluate the quality of the induced labels, we compare our method with two baselines. The first is BERTopic \cite{grootendorst2022bertopic}. We use the same underlying clusters for both approaches and provide the same LLM with the cluster representative documents; for the BERTopic condition, the model additionally receives the ten highest-ranked cluster keywords.

The second baseline is inspired by TopicGPT~\cite{pham2024topicgpt}. For each cluster, the LLM receives the representative advertisements and generates a concise topic label independently, without access to previously generated labels or an evolving shared taxonomy. This comparison is intended to isolate the effect of iterative taxonomy construction relative to single-shot LLM labeling.

Two volunteer annotators (one female and one male) with backgrounds in natural language processing (NLP) and computational social science (CSS), age range 25-40, evaluate the resulting labels. Each annotator is given the same set of 10 advertisements. For each evaluated advertisement, the labels produced by BERTopic, the TopicGPT-style baseline, and our framework are presented in randomized order. Annotators select the label that best represents the advertisement and rate each candidate label from $1$ (irrelevant) to $5$ (perfect fit).

\subsection{Moral Foundation Annotation}

We use Moral Foundations Theory~\cite{iyer2012understanding,haidt2007morality,haidt2004intuitive} as an exploratory framework for examining the moral framing of political advertisements. We consider six foundations: Care/Harm, Fairness/Cheating, Loyalty/Betrayal, Authority/Subversion, Sanctity/Degradation, and Liberty/Oppression. For more information on Moral Foundations Theory, see App.~\ref{sec:moral_foundation}.

For each topic-labeled cluster, we first prompt the LLM to summarize the primary argument expressed by its representative advertisements with respect to the assigned topic. We then provide the model with definitions of the six moral foundations and use constrained decoding to require selection of a single foundation. The complete prompts are provided in App.~\ref{app:prompts}.

Because these moral labels are generated using the topic annotation as part of the prompting procedure, we treat subsequent relationships between topics and moral foundations as exploratory downstream analyses rather than independent validation of the topic taxonomy.

\subsection{Document-Level Classification}

The taxonomy-induction framework produces cluster-level labels, while HDBSCAN leaves a substantial fraction of advertisements unassigned. To examine whether the induced topic labels can be propagated to individual advertisements, we conduct a separate supervised classification experiment.

We construct a pseudo-labeled training set from advertisements assigned to clusters with HDBSCAN membership probability greater than $0.98$. This produces $2{,}337$ advertisements. The resulting class distribution is highly imbalanced: the four largest classes---\textit{voting rights} ($26\%$), \textit{crime/justice} ($19\%$), \textit{education} ($17\%$), and \textit{abortion} ($11\%$)---account for $73\%$ of the pseudo-labeled advertisements, while the smallest classes, \textit{healthcare access} and \textit{property taxes}, each account for $0.64\%$.

We additionally sample $200$ advertisements from among the unassigned documents and annotate them manually with topic labels. These annotations serve as human-labeled evaluation data for the document-level classification task.

We compare Logistic Regression, XGBoost~\cite{chen2016xgboost}, RoBERTa~\cite{liu2019roberta}, and SetFit~\cite{tunstall2022efficient}. For classifiers operating on fixed embeddings, advertisement texts are represented using \texttt{all-MiniLM-L6-v2}. Hyperparameters are selected using GridSearchCV, and performance is reported using macro F1 and accuracy. The manually annotated advertisements were held out entirely for final evaluation.
\begin{table}[t]
    \centering
    \small
    \begin{tabular}{|c|c|c|}
        \hline
        \textbf{Model} & \textbf{Average Score} & \textbf{Best Label} \\ \hline
        BERTopic & 1.2 & 3 \\ \hline
        TopicGPT-style & 2.7 &  5\\ \hline
        Our Method & 2.8 & 12 \\ \hline
    \end{tabular}
    \caption{{\small Annotation results. The average score is out of 5, and the best label is the number of times that model's label was selected as the most fitting (among all annotators).}}
    \label{tab:annotation-results}
\end{table}

\subsection{Ad-Level Annotation}
For downstream descriptive analyses, we additionally assign topic and moral-foundation labels to individual advertisements using the induced taxonomy. We use \texttt{mistralai/Mistral-7B-v0.1} \cite{jiang2023mistral} for this ad-level annotation stage. This model is used only for the downstream case-study analyses and is separate from the Llama-3.2-3B model used to induce and evaluate the cluster-level taxonomy.

Using these document-level annotations together with Meta's spend and impression data, we examine issue prevalence, advertising expenditure, funding concentration, moral framing, and demographic exposure patterns.

To characterize associations between advertisement content and demographic groups, we compute positive pointwise mutual information (PPMI)~\cite{church1990word} between topic or moral-foundation annotations and observed audience attributes. Higher PPMI indicates that a topic or moral frame occurs disproportionately among advertisements delivered to a given demographic group. We interpret these associations descriptively as exposure patterns rather than direct evidence of advertisers' targeting intent.
\begin{table}[t]
    \centering
    \small
    \begin{tabular}{|l |c|c|}
        \hline
        \textbf{Model} & \textbf{Macro F1} & \textbf{Accuracy} \\
        \hline
        Logistic Regression & {0.37} & 0.55 \\
        \hline
        XGBoost             & 0.31 & 0.53 \\
        \hline
        RoBERTa             & 0.32 & 0.53 \\
        \hline
        SetFit              & 0.36 & {0.60} \\
        \hline
    \end{tabular}
    \caption{{\small Document-level classification performance for topic assignment.}}
    \label{tab:classification_accuracy}
\end{table}
\section{Results and Analysis}
\label{sec:res}

We first evaluate the quality of the induced topic labels and the feasibility of propagating those labels to individual advertisements. We use the resulting annotations for exploratory analyses of issue salience, moral framing, advertising expenditure, funding concentration, and demographic exposure.
\begin{figure}
    \centering

    \begin{minipage}[c]{0.88\columnwidth}
        \centering

        \begin{subfigure}{\linewidth}
            \centering
            \includegraphics[width=\linewidth]{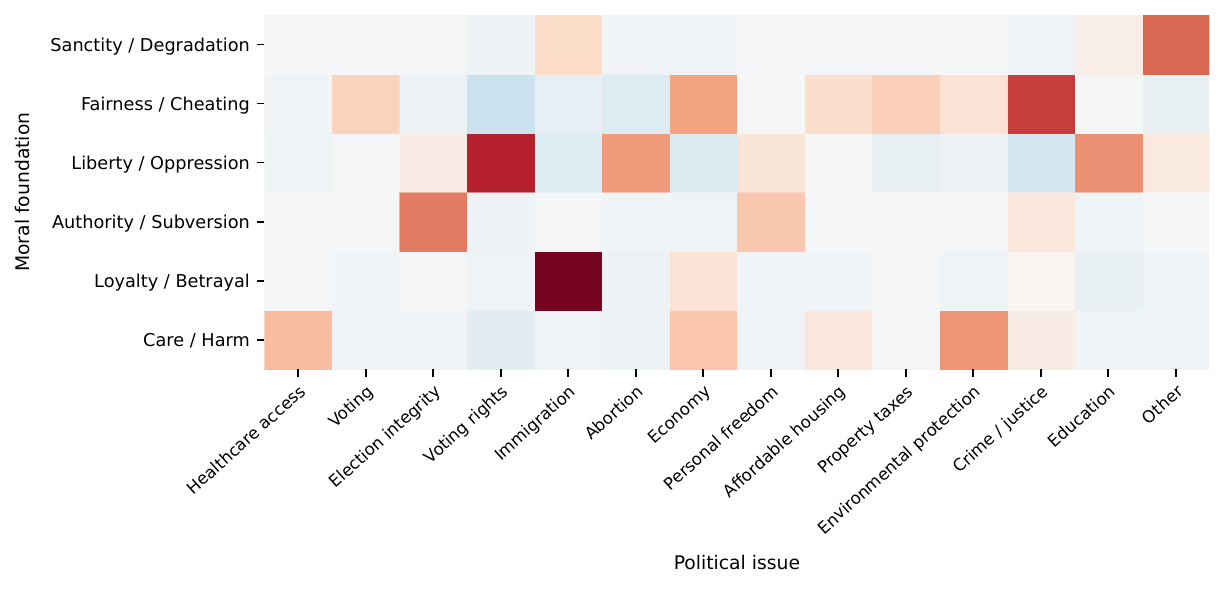}
            \caption{Proposed Framework Topics}
            \label{fig:moral_foundation_vs_topics-proposed}
        \end{subfigure}

        \begin{subfigure}{\linewidth}
            \centering
            \includegraphics[width=\linewidth]{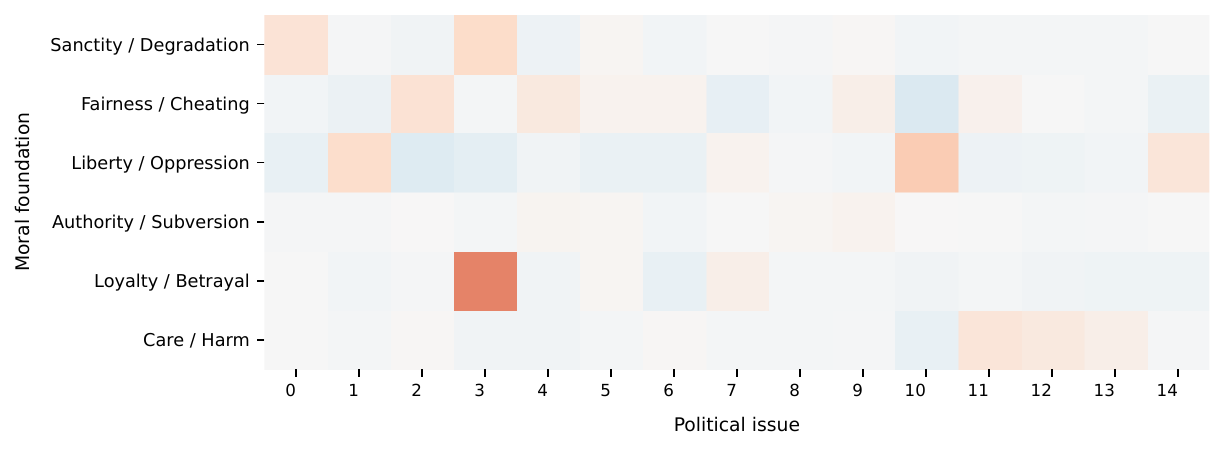}
            \caption{LDA Topics}
            \label{fig:moral_foundation_vs_topics-lda}
        \end{subfigure}

    \end{minipage}
    \hfill
    \begin{minipage}[c]{0.10\columnwidth}
        \centering
        \includegraphics[
            width=\linewidth,
            keepaspectratio
        ]{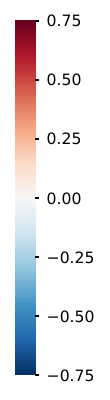}
    \end{minipage}

    \caption{\small{Associations between moral-foundation labels and topic labels for
    (a) the proposed taxonomy and (b) LDA topics. Cells show Pearson correlations
    between one-hot topic and moral-foundation assignments.}}
    \label{fig:moral_foundation_vs_topics}
\end{figure}
\subsection{Topic Label Evaluation}

Table~\ref{tab:annotation-results} summarizes the human evaluation of topic labels generated by BERTopic, the TopicGPT-style single-shot baseline, and our iterative framework.

The iterative framework receives the highest average rating ($2.8$), compared with $2.7$ for the single-shot baseline and $1.2$ for BERTopic. The difference in mean rating between our method and the single-shot baseline is small. However, the labels produced by our method are selected as the best-fitting label in $12$ judgments, compared with $5$ for the single-shot baseline and $3$ for BERTopic.

These results suggest that maintaining a shared, evolving topic inventory can improve label consistency relative to generating cluster labels independently, although the small scale of the human evaluation limits the strength of this conclusion. The strongest contrast is with BERTopic-based labeling, while the difference between iterative and single-shot LLM labeling is more modest in average rating.

The annotators achieve a Cohen's $\kappa$ \citep{cohen1960coefficient} of $0.66$, indicating substantial agreement under this evaluation setup. For $5$ out of the $20$ rounds, the annotators gave the same score for all models (in all cases, a score of 1). 
\begin{figure}[t]
    \centering
    \includegraphics[width=\linewidth]{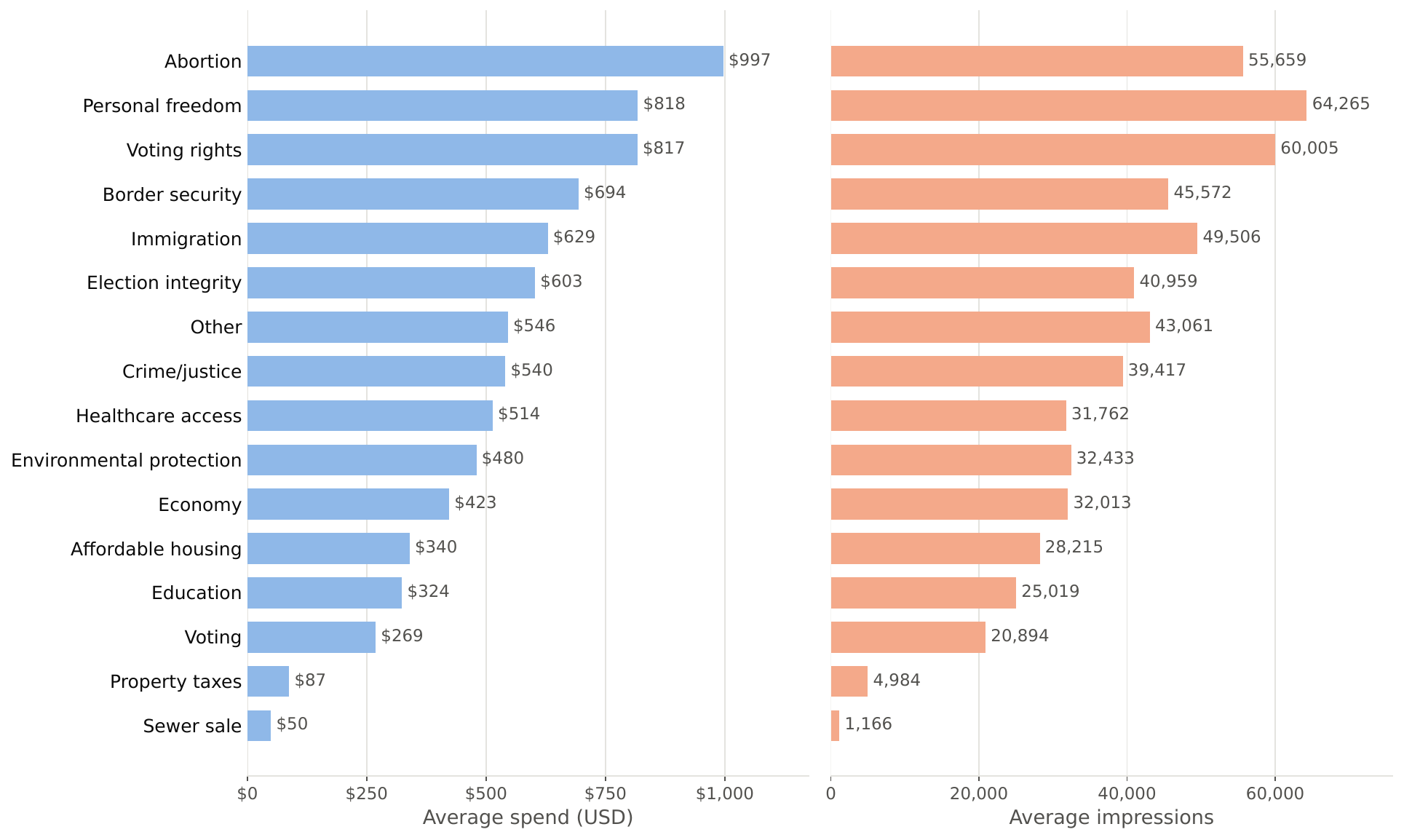}
    \caption{{\small Average advertising spend and impressions per ad by annotated topic.}}
    \label{fig:topic_spend_view}
\end{figure}
\subsection{Document-Level Classification Results}
We next evaluate whether the cluster-derived labels can be propagated to individual advertisements. Table~\ref{tab:classification_accuracy} reports performance across the four supervised classifiers. Logistic Regression achieves the highest macro F1 score ($0.37$), while SetFit achieves the highest overall accuracy ($0.60$). Performance is modest across all models, reflecting the difficulty of the task and the substantial class imbalance in the pseudo-labeled training data.

These results indicate that the induced labels can provide supervision for document-level classification, but they also show that propagating cluster-level topic labels to individual advertisements remains challenging. We therefore treat the supervised classifier as a downstream extension of the framework rather than evidence of high-quality annotation for all advertisements.
\begin{figure*}
    \centering
    \includegraphics[width=\linewidth]{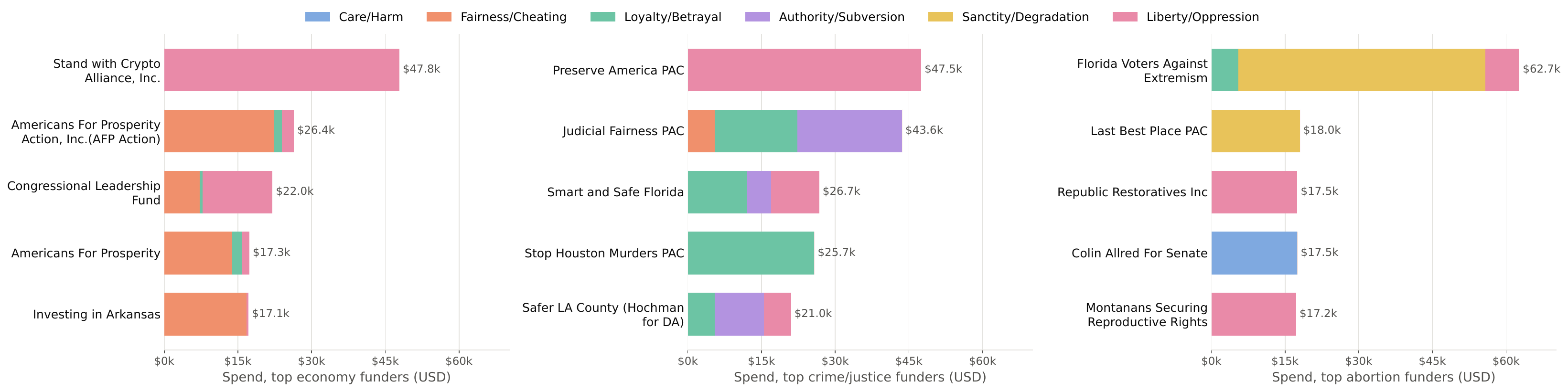}
    \caption{{\small Spend by the top five funders of economy, crime/justice, and abortion ads, segmented by moral foundation.}}
    \label{fig:top5_fe_MF_spend_Economy_Abortion}
\end{figure*}
\subsection{Topic Salience and Moral Framing}
We next examine how the induced topics relate to the moral-foundation annotations. For each advertisement, we construct one-hot representations of its topic and moral-foundation labels and compute pairwise Pearson correlations.

Figure~\ref{fig:moral_foundation_vs_topics} compares these relationships for the induced taxonomy and an LDA baseline. In the induced taxonomy, several topic--foundation pairs show comparatively strong associations. For example, \textit{Fairness/Cheating} is strongly associated with \textit{crime/justice}, while \textit{Loyalty/Betrayal} is strongly associated with \textit{immigration}. The corresponding LDA topics show weaker associations with the moral-foundation labels.

Because the moral labels are themselves generated using topic-conditioned LLM inference, these relationships should be interpreted as descriptive properties of the resulting annotation pipeline rather than independent validation of the topic taxonomy. Nevertheless, they illustrate how an interpretable topic inventory can support subsequent analysis of moral framing.
\begin{figure}[t]
    \centering
    \includegraphics[width=\columnwidth]{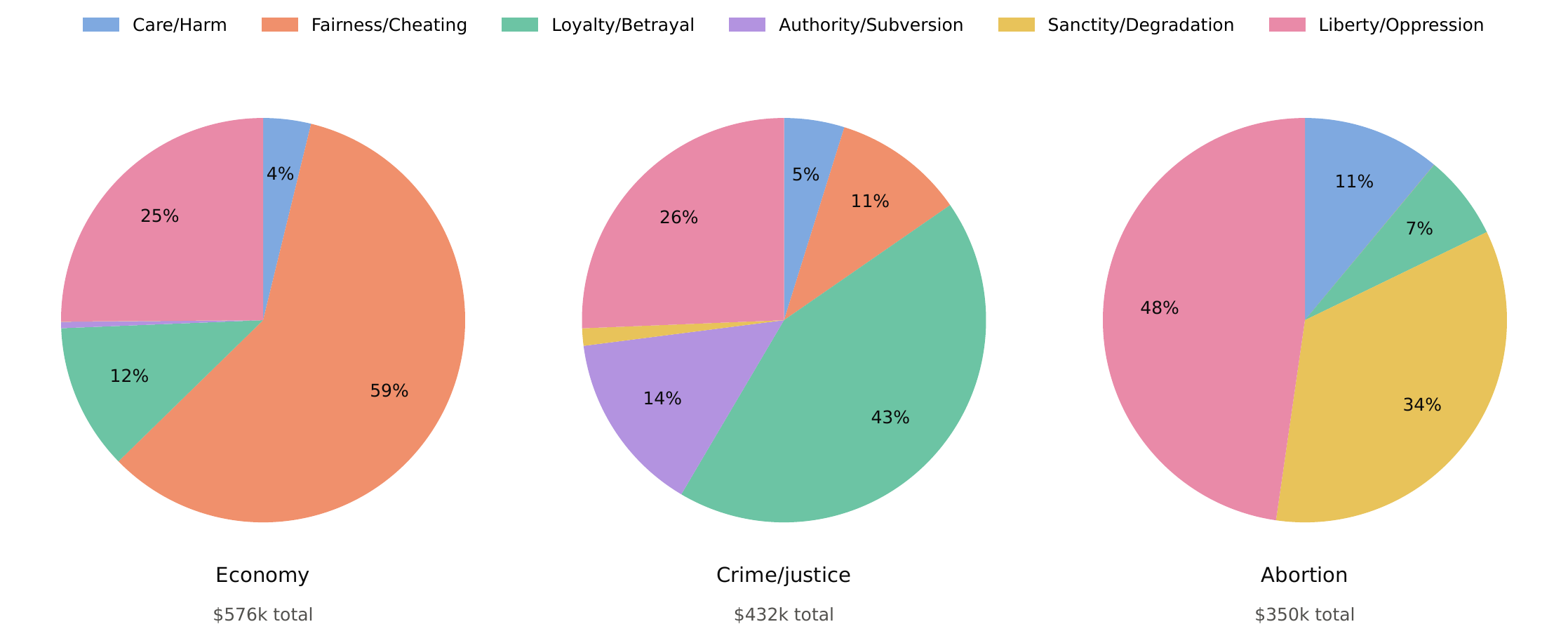}
    \caption{{\small Distribution of advertising spending across moral foundations within economy, crime/justice, and abortion advertisements. Percentages indicate within-topic spending shares; total spending is shown beneath each chart.}}
    \label{fig:ad_spend_by_MF}
\end{figure}
\subsection{Ad-Level Analysis}

We further use the document-level annotations to examine advertising expenditure, audience exposure, funding concentration, and moral framing. These analyses are descriptive and are intended to demonstrate the analytical affordances of the induced taxonomy rather than establish causal claims about campaign strategy.

\subsubsection{Spending and Impressions by Topic}

Figure~\ref{fig:topic_spend_view} compares average advertising spend and average impressions across topics. \textit{Abortion}, \textit{personal freedom}, and \textit{voting rights} have among the highest average spending levels in the annotated corpus. By contrast, \textit{immigration} and \textit{border security} receive comparatively high impressions relative to their average spending.

Lower-volume issues such as \textit{property taxes} and local-service topics receive substantially less expenditure and exposure. Overall, the differences between spending and impressions illustrate that issue categories vary not only in advertiser investment but also in the amount of observed audience exposure associated with that investment.

\subsubsection{Leading Funders by Issue}




Figure~\ref{fig:top5_fe_MF_spend_Economy_Abortion} compares estimated advertising expenditure among the five highest-spending funders within each of three issue categories: \textit{economy}, \textit{abortion}, and \textit{crime/justice}\footnote{\url{https://www.pewresearch.org/politics/2024/09/09/issues-and-the-2024-election/}} 
. For economy and abortion advertisements, the largest displayed funder spends substantially more than each of the other four. In contrast, crime/justice has two leading funders with similar expenditures, followed by three funders with lower, broadly comparable spending.

These results illustrate differences in expenditure patterns among the leading funders across issue categories in the observed corpus. Because the comparison is restricted to the five highest-spending funders in each category, it does not establish differences in overall funding concentration or in the total number of organizations supporting each issue. Assessing those questions requires the complete funder-level expenditure distribution. Further analysis is provided in Appendix~\ref{app:tf}.

\subsubsection{Moral Framing and Advertising Spend}
Figure~\ref{fig:ad_spend_by_MF} shows the distribution of estimated advertising expenditure across moral foundations within each issue category. Economic advertising primarily emphasizes \textit{Fairness/Cheating} (59\%) and \textit{Liberty/Oppression} (25\%), while crime/justice advertising primarily emphasizes \textit{Loyalty/Betrayal} (43\%) and \textit{Liberty/Oppression} (26\%). Abortion advertising primarily emphasizes \textit{Liberty/Oppression} (48\%) and \textit{Sanctity/Degradation} (34\%).

Figure~\ref{fig:top5_fe_MF_spend_Economy_Abortion} shows that the composition of expenditure also varies among the highest-spending funders within each category. For example, the displayed abortion funders include expenditure associated with \textit{Care/Harm}, in addition to the two largest aggregate foundations. These findings describe differences in spending associated with LLM-generated moral annotations; they do not independently establish funder motives or moral polarization. Further analysis is provided in Appendix~\ref{app:mf_fund}.

\subsubsection{Demographic Exposure Patterns}

Finally, we use PPMI to identify topic and moral-foundation labels that occur disproportionately among advertisements delivered to particular demographic groups.


\begin{figure*}[!t]
    \centering

    \begin{minipage}[c]{0.90\textwidth}
        \vspace{0pt}
        \centering

        \begin{subfigure}[b]{0.48\linewidth}
            \includegraphics[width=\linewidth]
                {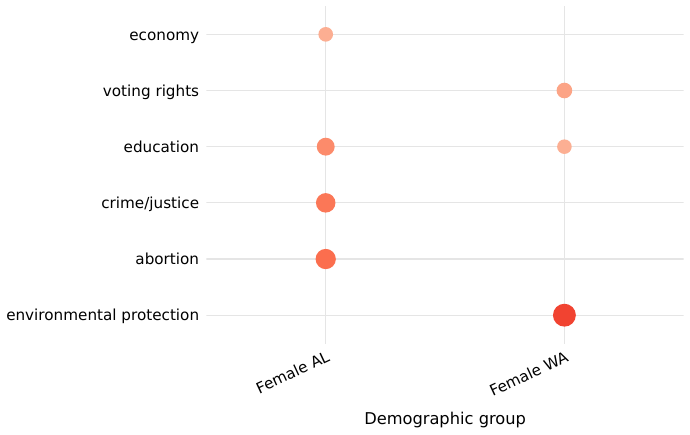}
            \caption{{\small Females: AL vs WA}}
        \end{subfigure}
        \hfill
        \begin{subfigure}[b]{0.48\linewidth}
            \includegraphics[width=\linewidth]
                {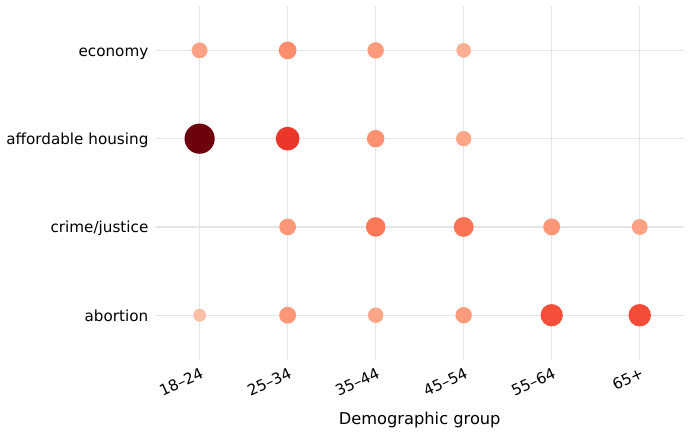}
            \caption{{\small All age groups: FL}}
            \label{fig:heatmap_2}
        \end{subfigure}

        \begin{subfigure}[b]{0.48\linewidth}
            \includegraphics[width=\linewidth]
                {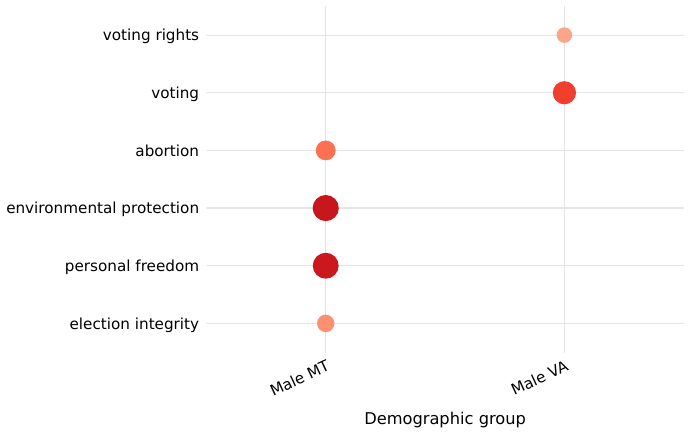}
            \caption{{\small Males: MT vs VA}}
            \label{fig:heatmap_3}
        \end{subfigure}
        \hfill
        \begin{subfigure}[b]{0.48\linewidth}
            \includegraphics[width=\linewidth]
                {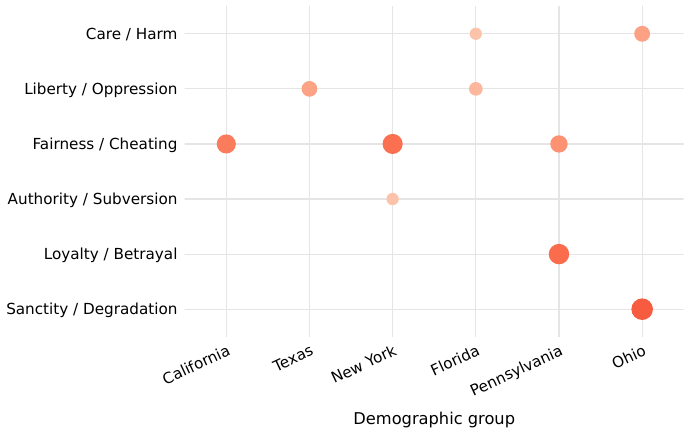}
            \caption{{\small Moral framing across selected states}}
        \end{subfigure}
    \end{minipage}
    \hfill
    \begin{minipage}[c]{0.06\textwidth}
        \vspace{0pt}
        \centering
        \includegraphics[
            width=\linewidth,
            height=0.52\textheight,
            keepaspectratio
        ]{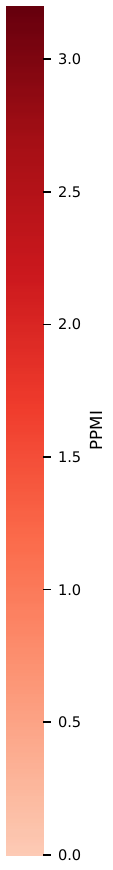}
    \end{minipage}

    \caption{{\small
    PPMI associations between political issue or moral-foundation labels
    and observed audience demographics. Higher values indicate that
    a label occurs disproportionately among advertisements delivered to
    the corresponding demographic group.
    }}
    \label{fig:heatmaps}
\end{figure*}

Figure~\ref{fig:heatmaps} shows several descriptive differences in exposure. In Florida, younger audiences receive a disproportionate share of \textit{affordable housing} advertisements, while older audiences show stronger associations with \textit{abortion}-related advertising. Geographic variation is also visible among male audiences: advertisements delivered to males in Montana show stronger associations with \textit{environmental protection} and \textit{personal freedom}, whereas advertisements delivered to males in Virginia show stronger associations with \textit{voting}.

These patterns demonstrate that issue exposure varies across demographic groups in the observed Meta delivery data. Because the dataset describes realized audience exposure rather than the full set of advertiser-selected targeting parameters, we \textbf{do not} interpret these associations as direct evidence of advertiser targeting intent.

\section{Conclusion}

We introduced a seed-free framework for inducing interpretable topic taxonomies from unlabeled text corpora. The framework combines embedding-based clustering with iterative LLM inference, maintaining a shared topic inventory that expands as new clusters are processed and is subsequently used to assign labels across the complete set of clusters.

We evaluated the framework through a case study of political advertisements from the month preceding the 2024 U.S. presidential election. In a small-scale human evaluation, the induced labels received similar average ratings to a single-shot LLM baseline but were selected as the best-fitting labels more often. We further showed that the resulting taxonomy can support downstream analyses of issue prevalence, moral framing, advertising expenditure, funding concentration, and demographic exposure.

The framework is not specific to political advertising and can be applied to other large, unlabeled text corpora in which interpretable topic organization is useful. Future work should evaluate the stability of induced taxonomies across domains, clustering configurations, LLMs, prompts, and cluster-processing orders, and explore methods for constructing more robust or hierarchical topic inventories.

\section{Limitations}


Our framework depends on the quality of the underlying embeddings and clustering procedure, and the resulting taxonomy may be sensitive to clustering parameters and the order in which clusters are processed. The human evaluation is also small in scale, and the difference in average rating between our framework and the single-shot LLM baseline is modest. Larger evaluations across datasets and models are needed to assess the robustness of the approach.

The framework also relies on LLM-generated labels, which may introduce bias, oversimplify ambiguous texts, or produce misleading interpretations \cite{islam-2026-gets,ji2023survey,blodgett2020language,brown2020language}. The moral-foundation labels are generated using topic-conditioned inference and should therefore be interpreted as exploratory rather than as independent validation of the induced topics.

Finally, the case study covers approximately one month of Meta political advertising during the 2024 U.S. election cycle. Its substantive findings may not generalize to other periods, platforms, or political contexts. Meta reports spending and impressions as ranges, and its demographic statistics describe observed ad delivery rather than necessarily reflecting advertiser targeting intent.

\section{Ethical Considerations}

This study uses political advertising data made publicly available through the Meta Ad Library. The data contain aggregate audience statistics and do not include personally identifiable information.

As the topic and moral annotations are partly generated by LLMs, they may reflect model biases or simplify nuanced political discourse. We therefore interpret the resulting patterns descriptively and avoid treating demographic exposure patterns as evidence of individual political preferences or advertiser intent. We report the models, prompts, and processing details used in our experiments to support transparency and reproducibility.


\section{Acknowledgments} 
We would like to thank Lightning AI Studio for providing compute resources. We also thank the anonymous reviewers for their thoughtful feedback. 
\bibliography{custom}

\appendix


\section{Prompts}
\label{app:prompts}
Below are the prompts used in the various steps of the framework. Constrained decoding was implemented with the guidance framework\footnote{https://github.com/guidance-ai/guidance}.
\smallskip

\textbf{System Prompt}
\begin{quote}
You are a political analyst. You are given the following set of political ads to analyze. You will assign the ads a topic that best summarizes the key issue discussed in these ads. 
\end{quote}
\smallskip

\subsection{Topic Synthesis}

\textbf{Binary Output}
\begin{quote}
Is there a topic in the following list that well describes the key issue discussed in these ads? 

    Topics:
    
    \{\{ topics \}\}

Can the ads be summarized by one of the previous topics? Answer with "yes" or "no".
\end{quote}
\medskip

\textbf{Generate New Topic (User)}
\begin{quote}
    In three words or less, describe the issue that ads discuss, summarizing the topic. Examples: "Abortion", "Climate Change", etc.
\end{quote}
\medskip

\textbf{Generate New Topic (Assistant)}
\begin{quote}
    A better topic for these ads is: "
\end{quote}
The LLM would finish the above, with the quote (") as a stop token.
\medskip

\subsection{Annotation}

\textbf{System Prompt}
\begin{quote}
You are a political analyst. You are given the following set of political ads to analyze: \{\{ ads \}\}
\end{quote}
\medskip

\textbf{Select Topic (User)}
\begin{quote}
Which of the following topics best describes the key issue discussed in these ads? Select exactly one topic from the provided list.

 Topics:
    
    \{\{ topics \}\}
\end{quote}
\medskip

\textbf{Summarize (User)}
\begin{quote}
Summarize the main talking point of the ads. Do so in a single sentence.
\end{quote}
\medskip

\textbf{Select Moral Foundation (User)}
\begin{quote}
Which of the following moral foundations best describes the arguments used in the ads about  \{\{ topic \}\}?

\{\{ MFs\_with\_definitions \}\}
\end{quote}

\section{Moral Foundations Theory}
\label{sec:moral_foundation}

Moral Foundations Theory (MFT)~\citep{iyer2012understanding,haidt2007morality,haidt2004intuitive} suggests a theoretical framework for analyzing \textit{six} moral values (i.e., foundations, each with a positive and a negative polarity) central to human moral sentiment. MFT states that political attitudes are shaped by \textit{six} core moral dimensions: care/harm, fairness/cheating, loyalty/betrayal, authority/subversion, sanctity/degradation, and liberty/oppression.

The definitions given in Table~\ref{tab:moral_foundations} correspond to the same definitions given to the LLM when assigning moral foundations. 

\begin{table*}
    \begin{center}
    \scalebox{.85}{\begin{tabular}{>{\arraybackslash}m{18cm}}
    \toprule
    \textsc{\textbf{Care/Harm:}} It suggests that someone other than the speaker is worthy of compassion or is experiencing harm, grounded in the values of kindness, tenderness, and care. \\
    \hline
    \textsc{\textbf{Fairness/Cheating:}} Emphasizes justice, personal rights, and independence; involves comparing with other groups. Advocates for equal opportunity and resists those who benefit without contributing (``Free Riders"). \\
    \hline
    \textsc{\textbf{Loyalty/Betrayal:}} Based on the values of loyalty to one’s country and willingness to sacrifice for the group. Activated by a sense of unity and collective responsibility—``one for all, and all for one".\\
    \hline
    \textsc{\textbf{Authority/Subversion:}}  Centers on showing respect (or resistance) toward established authority and following long-standing traditions. It includes maintaining social order and fulfilling the duties tied to hierarchical roles, such as obedience, respect, and role-based responsibilities.\\
    \hline
    \textsc{\textbf{Sanctity/Degradation:}} Beyond religion, this value highlights respect for human dignity and aversion to moral or physical corruption, promoting purity, self-control, and the belief that the body is sacred and vulnerable to defilement.\\
    \hline 
    \textsc{\textbf{Liberty/Oppression:}} Captures the feelings of reactance and resentment people experience when their freedom is restricted, often leading to collective disdain for authoritarian figures and motivating unity and resistance against oppression. \\
    \bottomrule
    \end{tabular}}
    \caption{{  Six basic moral foundations \cite{iyer2012understanding,haidt2007morality,haidt2004intuitive}.}}
    \label{tab:moral_foundations}
    \end{center}
\end{table*}

\section{Substantive Insights: Topic Salience and Moral Framing}
\label{app:mft}
We analyze the distribution of moral framings and political issues across different demographics to uncover political microtargeting strategies. We used the positive pointwise mutual information (PPMI) to detect topics that appear disproportionately in ads shown to specific demographics, potentially indicating targeted messaging. We visualize the results of the PPMI analysis in Figure~\ref{fig:heatmaps}. The plots show the distribution of moral framings and political issues across different demographics. Some interesting observations include the differences in ads targeting different states, or the shift in topics among different age groups in Florida (Figure \ref{fig:heatmap_2}). For example, \textit{{young}} people in {Florida} are more likely to be targeted with ads related to \textit{affordable housing}, while {\textit{older}} people are more likely to be targeted with ads related to \textit{abortion}. This shows generational targeting strategies, where {\texttt{economic concerns}} are emphasized for {younger groups}, while {older groups} are mobilized on {\texttt{moralized issues}} like abortion.

On the other hand, Figure \ref{fig:heatmap_3} shows that \textit{environmental protection} and \textit{personal freedom} are heavily overrepresented among {\textit{male}} audiences in {Montana}, while \textit{abortion} also appears moderately elevated. By contrast, {\textit{male}} audiences in {Virginia} are disproportionately exposed to \textit{voting}-related advertising. These contrasts highlight how campaign strategies selectively target localized demographic niches: {\texttt{progressive-leaning issue frames resonate more strongly in Montana male}} audiences, whereas mobilization around {\texttt{law-and-order and electoral process is more pronounced among Virginia males}}.

\section{Substantive Insights: Ad-Level Analysis}
\label{app:ad}
Using the synthesized topics, we annotate individual ads to enable ad-level analysis. Cluster-level annotations reveal latent issue and moral structure, while ad-level annotations enable analysis of spending, funder concentration, and targeting strategies.
\subsection{Top Funders}
\label{app:tf}
We examine the concentration of advertising spend among funding entities across major issue domains.
Figure~\ref{fig:top5_fe_MF_spend_Economy_Abortion} presents the top $5$ funding entities by total spend for three central issues in the 2024 election—\textit{{economy}}, \textit{{abortion}}, and \textit{{crime/justice}}.

For the \textit{economy}, Figure~\ref{fig:top5_fe_MF_spend_Economy_Abortion} shows substantial variation in advertising expenditure among the five highest-spending funders in the observed corpus. Stand with Crypto Alliance, Inc.\ is the largest displayed spender, at approximately \$47,800, followed by Americans for Prosperity Action, Inc.\ (\$26,400) and the Congressional Leadership Fund (\$22,000). Americans for Prosperity and Investing in Arkansas spend approximately \$17,300 and \$17,100, respectively. These results describe spending differences among the leading funders; assessing concentration across all economy-related advertising requires accounting for expenditure by the remaining funding entities.

For \textit{abortion}, Figure~\ref{fig:top5_fe_MF_spend_Economy_Abortion} shows a substantial spending gap among the five highest-spending funders in the observed corpus. Florida Voters Against Extremism\footnote{\href{https://dos.elections.myflorida.com/committees/ComDetail.asp?account=84315}{https://dos.elections.myflorida.com/committees}} is the largest displayed spender, with estimated expenditure of approximately \$62,700. The other four funders spend comparatively similar amounts: Last Best Place PAC (\$18,000), Republic Restoratives Inc.\ (\$17,500), Colin Allred for Senate (\$17,500), and Montanans Securing Reproductive Rights\footnote{\url{https://www.influencewatch.org/organization/montanans-securing-reproductive-rights/}} (\$17,200). These results highlight differences in expenditure among the leading funders, with a pronounced drop from the largest spender to the remaining four. Assessing concentration across all abortion-related advertising requires accounting for expenditure by the remaining funding entities.

In contrast, for \textit{crime/justice}, the two highest-spending funders have similar estimated expenditures: Preserve America PAC (\$47,500) and Judicial Fairness PAC (\$43,600). The remaining displayed funders---Smart and Safe Florida (\$26,700), Stop Houston Murders PAC (\$25,700), and Safer LA County (Hochman for DA) (\$21,000)---have lower, broadly comparable expenditures. These amounts show a gap between the leading pair and the other three funders within the displayed group.

Overall, Figure~\ref{fig:top5_fe_MF_spend_Economy_Abortion} illustrates differences in expenditure patterns among the five highest-spending funders within each issue category. For \textit{economy} and \textit{abortion}, the largest displayed funder spends substantially more than each of the other four, whereas \textit{crime/justice} has two leading funders with similar expenditures. These comparisons characterize spending differences among the leading funders in the observed corpus. They do not establish whether overall spending is more concentrated in one issue category or whether an issue involves a broader set of funding entities; evaluating those questions requires the complete funder-level expenditure distribution.

\subsection{Moral Foundations in Funded Messaging}
\label{app:mf_fund}
We examine the relationship between annotated moral framing and estimated advertising expenditure across three issue categories: \textit{economy}, \textit{crime/justice}, and \textit{abortion}. Figure~\ref{fig:ad_spend_by_MF} presents the distribution of spending across moral foundations within each topic, together with the corresponding total expenditure. This aggregate view summarizes the moral-foundation composition of spending in the analyzed corpus without disaggregating it by funding entity.

Figure~\ref{fig:top5_fe_MF_spend_Economy_Abortion} complements this aggregate view by comparing the five highest-spending funders within each of the three issue categories. Each funder's expenditure is segmented according to the moral-foundation labels assigned to its advertisements. These funder-level comparisons describe the displayed organizations rather than the complete population of funders.

For the \textit{economy}, \textit{Fairness/Cheating} accounts for approximately 59\% of topic-level expenditure, followed by \textit{Liberty/Oppression} at 25\%. The displayed funders differ in the composition of their spending. Stand with Crypto Alliance, Inc.\ has expenditure predominantly associated with \textit{Liberty/Oppression}-annotated advertisements. In contrast, Americans for Prosperity Action, Inc., Americans for Prosperity, and Investing in Arkansas have expenditure primarily associated with \textit{Fairness/Cheating}. The Congressional Leadership Fund's expenditure includes both \textit{Fairness/Cheating} and \textit{Liberty/Oppression}, while smaller \textit{Loyalty/Betrayal} components appear for some displayed funders. These patterns show variation in the annotated moral framing associated with expenditure among the leading economy-related funders.

For \textit{crime/justice}, the largest aggregate spending shares are associated with \textit{Loyalty/Betrayal} (43\%) and \textit{Liberty/Oppression} (26\%). The funder-level breakdown reveals different compositions within this issue category. Preserve America PAC's expenditure is predominantly associated with \textit{Liberty/Oppression}, whereas Stop Houston Murders PAC's expenditure is predominantly associated with \textit{Loyalty/Betrayal}. Judicial Fairness PAC has substantial expenditure associated with both \textit{Authority/Subversion} and \textit{Loyalty/Betrayal}. Thus, the aggregate distribution does not imply that all displayed funders exhibit the same moral-framing profile.

For \textit{abortion}, \textit{Liberty/Oppression} and \textit{Sanctity/Degradation} account for approximately 48\% and 34\% of topic-level expenditure, respectively. Among the displayed funders, Florida Voters Against Extremism and Last Best Place PAC have expenditure primarily associated with \textit{Sanctity/Degradation}. Republic Restoratives Inc.\ and Montanans Securing Reproductive Rights have expenditure predominantly associated with \textit{Liberty/Oppression}. However, Colin Allred for Senate's displayed expenditure is associated with \textit{Care/Harm}. The funder-level results therefore include additional moral foundations beyond the two largest aggregate categories and should not be reduced to a uniform split between liberty-based and sanctity-based framing.

Taken together, Figures \ref{fig:top5_fe_MF_spend_Economy_Abortion} and \ref{fig:ad_spend_by_MF}  show differences in the moral-foundation composition of advertising expenditure across issue categories and among their highest-spending funders. These findings describe spending associated with LLM-generated annotations; they do not independently establish funder motives, specific policy arguments, or moral polarization. We therefore interpret them as exploratory patterns in the annotated corpus rather than as independently validated accounts of the organizations' moral or political positions.

\end{document}

%% file: figure1.tex
\definecolor{Ink}{HTML}{090D20}
\definecolor{BlueFill}{HTML}{CEE4F8}
\definecolor{BlueLine}{HTML}{22658B}
\definecolor{GreenFill}{HTML}{DEEDD9}
\definecolor{GreenLine}{HTML}{407C46}
\definecolor{OrangeFill}{HTML}{FFDAC0}
\definecolor{OrangeLine}{HTML}{9C5327}
\definecolor{PurpleFill}{HTML}{E1C8F5}
\definecolor{PurpleLine}{HTML}{8640AE}
\definecolor{BoxLine}{HTML}{9396A0}
\newcommand{\Doc}[3]{%
  \begin{scope}[shift={(#1,#2)}]
    \path[draw=#3,fill=white,line width=1.5pt,rounded corners=3.5pt]
      (-20,-26) rectangle (20,26);
    \draw[#3,line width=1.4pt] (-9,-12)--(8,-12)
      (-9,-1)--(8,-1) (-9,10)--(5,10);
  \end{scope}}
\newcommand{\Cluster}[5]{%
  \begin{scope}[shift={(#1,#2)}]
    \path[shade,left color=#3,right color=#3!80!white,
      draw=none] (0,0) circle[radius=88];
    \ifnum#5=3
      \Doc{8}{-31}{#4}\Doc{-37}{13}{#4}\Doc{24}{34}{#4}
    \else
      \Doc{5}{-31}{#4}\Doc{-39}{12}{#4}\Doc{44}{12}{#4}\Doc{2}{48}{#4}
    \fi
  \end{scope}}
\newcommand{\Robot}[2]{%
  \begin{scope}[shift={(#1,#2)}]
    \draw[Ink,line width=1.6pt] (0,-28)--(0,-42);
    \draw[Ink,fill=white,line width=1.4pt] (0,-48) circle[radius=7];
    \path[draw=Ink,fill=BoxLine!25,line width=1.2pt,rounded corners=2pt]
      (-41,-9) rectangle (-32,14) (32,-9) rectangle (41,14);
    \path[draw=Ink,fill=white,line width=1.4pt,rounded corners=8pt]
      (-35,-28) rectangle (35,29);
    \fill[Ink] (-12,0) circle[radius=4.7];
    \fill[Ink] (12,0) circle[radius=4.7];
  \end{scope}}
\newcommand{\Bubble}[5]{%
  \begin{scope}[shift={(#1,#2)}]
    \path[draw=none,fill=#4,rounded corners=7pt]
      (0,0)--(#3,0)--(#3,160)--(133,160)
      [rounded corners=0pt]--(101,188)--(105,160)
      [rounded corners=7pt]--(0,160)--cycle;
    \node[anchor=north west,align=left,inner sep=0pt,
      font=\fontsize{12}{14}\selectfont] at (20,16) {#5};
  \end{scope}}
\newcommand{\Taxonomy}[4]{%
  \begin{scope}[shift={(#1,#2)}]
    \ifnum#4=1 \def\boxheight{190}\else\def\boxheight{128}\fi
    \path[draw=BoxLine,fill=gray!2,line width=.65pt,rounded corners=6pt]
      (0,0) rectangle (#3,\boxheight);
    \node[font=\bfseries\fontsize{12}{14}\selectfont] at (#3/2,32)
      {Current taxonomy};
    \path[fill=BlueFill,rounded corners=4.5pt]
      (23,58) rectangle (#3-23,107);
    \node at (#3/2,83) {Economy};
    \ifnum#4=1
      \path[fill=OrangeFill,rounded corners=4.5pt]
        (23,116) rectangle (#3-23,168);
      \node at (#3/2,142) {Education};
    \fi
  \end{scope}}

\begin{tikzpicture}[x=.015cm,y=-.015cm,text=Ink,
  font=\fontsize{12}{14}\selectfont,
  flow/.style={draw=Ink,line width=1.35pt,
    -{Stealth[length=5pt,width=5pt]},line join=round}]
  \path[use as bounding box] (25,70) rectangle (1210,1240);
  \node[anchor=west,font=\bfseries\fontsize{20}{23}\selectfont]
    at (38,109) {(a)\quad Iteratively build a shared topic taxonomy};

  \draw[flow] (217,405)--(268,405);
  \draw[flow] (368,405)--(441,405);
  \draw[flow] (619,405)--(654,405);
  \draw[flow] (747,405)--(824,405);
  \draw[flow] (1001,405)--(1035,405);
  \Cluster{128}{379}{BlueFill}{BlueLine}{4}
  \Cluster{532}{379}{GreenFill}{GreenLine}{3}
  \Cluster{913}{379}{OrangeFill}{OrangeLine}{4}
  \node at (128,486) {Cluster 1};
  \node at (532,486) {Cluster 2};
  \node at (913,486) {Cluster 3};
  \Robot{315}{405}\Robot{701}{405}\Robot{1080}{405}
  \node[font=\fontsize{19}{20}\selectfont] at (1170,401) {$\cdots$};
  \Bubble{239}{166}{207}{BlueFill}{No existing\\topic matches.\\Add new topic:\\\textbf{Economy.}}
  \Bubble{613}{166}{220}{GreenFill}{Matches existing\\topic (Economy).\\\textbf{No new topic}\\\textbf{added.}}
  \Bubble{1003}{166}{200}{OrangeFill}{No existing\\topic matches.\\Add new topic:\\\textbf{Education.}}
  \draw[flow] (315,446)--(315,523);
  \draw[flow] (701,446)--(701,523);
  \draw[flow] (1080,446)--(1080,523);
  \Taxonomy{159}{532}{302}{0}
  \Taxonomy{547}{532}{305}{0}
  \Taxonomy{926}{532}{277}{1}

  \node[anchor=west,font=\bfseries\fontsize{20}{23}\selectfont]
    at (38,842) {(b)\quad Assign topic labels to all clusters};

  \path[draw=BoxLine,fill=gray!2,line width=.65pt,rounded corners=6pt]
    (56,898) rectangle (348,1225);
  \node[font=\bfseries\fontsize{11.5}{14}\selectfont] at (202,930)
    {Final topic taxonomy};
  \foreach \yy/\col/\label in {953/BlueFill/Economy,
    1007/GreenFill/Immigration,1061/OrangeFill/Education,
    1117/PurpleFill/Voting}{
    \path[fill=\col,rounded corners=4.5pt]
      (80,\yy) rectangle (325,\yy+48);
    \node[anchor=west,inner sep=0pt] at (105,\yy+24) {\label};
  }
  \node[anchor=west,font=\bfseries\fontsize{22}{22}\selectfont]
    at (94,1186) {$\cdots$};

  \draw[flow] (348,1060)--(377,1060);
  \draw[Ink,line width=1.35pt,line join=round]
    (470,927)--(1025,927);
  \foreach \xx in {470,657,845,1025}
    \draw[flow] (\xx,927)--(\xx,970);
  \Cluster{470}{1060}{BlueFill}{BlueLine}{4}
  \Cluster{659}{1060}{GreenFill}{GreenLine}{4}
  \Cluster{844}{1060}{OrangeFill}{OrangeLine}{4}
  \Cluster{1033}{1060}{PurpleFill}{PurpleLine}{4}
  \foreach \xx/\col/\label in {381/BlueFill/Economy,
    571/GreenFill/Immigration,762/OrangeFill/Education,
    951/PurpleFill/Voting}{
    \path[draw=none,fill=\col,rounded corners=6pt]
      (\xx,1147) rectangle (\xx+180,1202);
    \node at (\xx+90,1174.5) {\label};
  }
  \node[font=\fontsize{19}{20}\selectfont] at (1160,1050) {$\cdots$};
\end{tikzpicture}